\documentclass[10pt,twocolumn,letterpaper]{article}

\usepackage{wacv}
\usepackage{times}
\usepackage{epsfig}
\usepackage{graphicx}
\usepackage{amsmath}
\usepackage{amssymb}
\usepackage{times}
\usepackage{subcaption}
\usepackage{amssymb}
\usepackage{gensymb}
\usepackage{stackengine}
\usepackage{tabu}
\usepackage{threeparttable}
\usepackage{soul}

\wacvfinalcopy 

\def\wacvPaperID{228} 

\ifwacvfinal\fi
\begin{document}

\title{360 Panorama Synthesis from a Sparse Set of Images \\with Unknown Field of View}

\author{Julius Surya Sumantri and In Kyu Park\\
{\tt\small \hspace*{4mm} \{julius.taeng@gmail.com \hspace*{0.5mm} pik@inha.ac.kr\}}\\
Dept. of Information and Communication Engineering, Inha University, Incheon 22212, Korea
\and
}

\maketitle
\ifwacvfinal\thispagestyle{empty}\fi

\begin{abstract}
360{\degree} images represent scenes captured in all possible viewing directions and enable viewers to navigate freely around the scene thereby providing an immersive experience.
Conversely, conventional images represent scenes in a single viewing direction with a small or limited field of view (FOV).
As a result, only certain parts of the scenes are observed, and valuable information about the surroundings is lost.
In this paper, a learning-based approach that reconstructs the scene in $360\degree\times180\degree$ from a sparse set of conventional images (typically 4 images) is proposed.
The proposed approach first estimates the FOV of input images relative to the panorama.
The estimated FOV is then used as the prior for synthesizing a high-resolution 360{\degree} panoramic output.
The proposed method overcomes the difficulty of learning-based approach in synthesizing high resolution images (up to 512$\times$1024).
Experimental results demonstrate that the proposed method produces 360{\degree} panorama with reasonable quality. Results also show that the proposed method outperforms the alternative method and can be generalized for non-panoramic scenes and images captured by a smartphone camera.
\end{abstract}
\vspace*{-5mm}
\section{Introduction}
Images are limited to the boundaries of what the camera can capture.
An image with a narrow field of view~(FOV) sees only a small part of a given scene.
After the scene is captured, the viewer obtains no information on what lies beyond the image boundary.
A 360{\degree} panoramic image overcomes this limitation through an unlimited FOV.
As a result, all information on scenes across the horizontal and vertical viewing directions are captured.
This imagery provides viewers with an outward-looking view of scenes and freedom to shift their viewing directions accordingly.
Furthermore, the images are widely used in many fields, such as virtual environment modeling, display system, free-viewpoint videos, and illumination modeling.

\begin{figure}[t]
	\begin{center}
		\includegraphics[width=\linewidth]{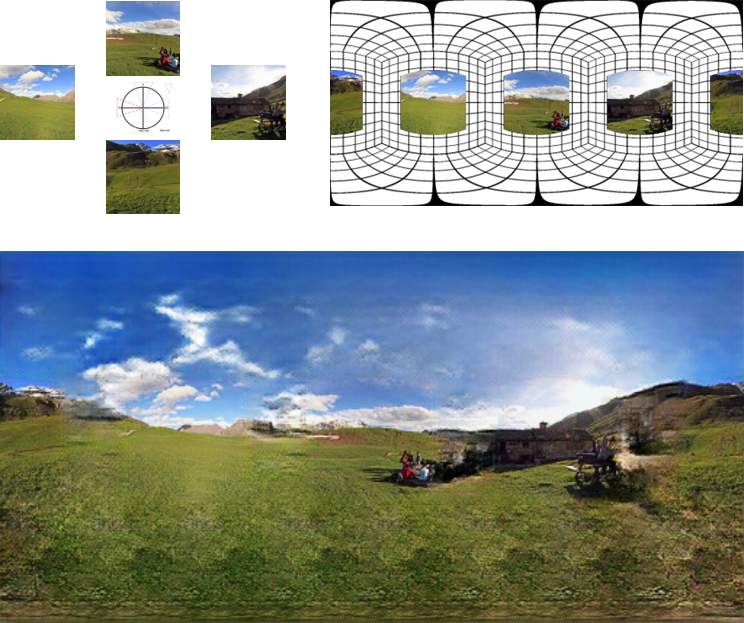}
	\end{center}
	\vspace*{-0.4cm}
	\caption{
		Overview of the proposed method to synthesize 360{\degree} panorama from partial input.
		The inputs are captured in 4 perpendicular and horizontal viewing directions without overlapping.
	}
	\label{fig:overview}
	\vspace*{-3mm}
\end{figure}

Generating a scene with a large FOV from a smaller one is a long-standing task in computer vision domain.
Given an image with small FOV, humans can easily expect what the image looks like in a larger FOV because of the human capability to estimate the scene outside of the viewing boundary~\cite{Intraub1989WideangleMO} that is learned during a lifetime.
However, for a computer vision task, an image with small FOV contains minimal information about the surrounding scene, making it an ill-posed problem and highly challenging task.

Conventional algorithms~\cite{Brown2007, Szeliski_2006} reconstruct a panoramic image with a large FOV by stitching multiple images and heavily rely on accurate homography estimation and feature matching over significantly overlapped regions of input images.
Therefore, these algorithms only synthesize a partial panorama or many images are needed to synthesize a full 360{\degree} panorama.

By contrast, our study aimed to solve the problem of synthesizing full 360{\degree} panoramic images from only a sequence of 4~images without any overlap.
These sequences are partial observations of scenes captured from 4 viewing directions as shown in Figure~\ref{fig:overview}.
Note that the camera focal length and FOV are assumed to be unknown.

Learning-based methods rooted in deep neural networks~(DNN) have achieved remarkable success and are widely used to solve various computer vision tasks~\cite{Ahn_2018_CVPR, Zolfaghari_2017_ICCV, Pathak_2016_CVPR, Soltani_2017_CVPR}.
The DNN encodes millions of parameters that are suitable for handling tasks that require complex data modeling.
In this study, A learning-based method is adopted to train the proposed model to significantly extrapolate and interpolate scenes to form full panoramic images.
The network learns the statistics of various general outdoor scenes to model their true distribution.

The proposed approach consists of two processing stages that are developed based on convolutional neural network (CNN) with generative adversarial framework.
The first stage is FOV estimation stage in which the main task is to predict input images' FOV relative to the full panoramic images.
This is an essential preprocessing step before the actual panoramic synthesis.
The estimated FOV from the input sequence is mapped into the panorama FOV, which we refer as relative FOV.
The second stage is the panorama synthesis, in which the output from the first stage is fed as the input to synthesize 360{\degree} $\times$ 180{\degree} panorama.

To the best of our knowledge, the proposed method is the first one to address the problem of relative FOV estimation and synthesis of 360{\degree} panoramic images from a sparse set of images without any overlap.
The contribution of this paper can be summarized as follows.
\begin{itemize}
	\item We proposed a model and network to estimate the relative FOV from a sparse set of images with an unknown FOV and no overlap {\bf (Sec. 3.1)}.
	\item We designed a novel deep neural network to synthesize full 360{\degree} $\times$ 180{\degree} panoramic images with high-resolution up to 512$\times$1024 {\bf (Sec. 3.2)}.
\end{itemize}


\section{Related Work}
\subsection{FOV Prediction}
DeepFocal~\cite{deepfocal2015} estimates the horizontal FOV of a single image using pre-trained features on AlexNet~\cite{NIPS2012_4824} architecture.
The network takes input of images pixels directly and finetuned to estimate the FOV.
It treated as a regression task by replacing the fully connected layers with a single node output.

\vspace{-1mm}
\subsection{Image Inpainting and Completion}
Inpainting methods interpolate missing or occluded regions of input by filling these regions with plausible pixels.
Most algorithms rely on neighboring pixels to propagate the pixel information to target regions~\cite{Ballester2001, bertalmio2000image}.
These methods generally handle images with narrow holes and do not perform well on large holes~\cite{Barnes2009, Kwatra2005}.
Liu \etal~\cite{Liu_2018_ECCV} recently proposed to solve arbitrary holes by training CNN.
To solve an inpainting task, a partial convolution with a binary mask is used as prior for the missing holes.

\vspace{-2mm}
\subsection{Novel View Synthesis}
This method generates images with different viewing directions.
The task includes generating different poses transformed by a limited rotation \cite{Tatarchenko2015}.
Dosovistsky~\cite{Dosovitskiy_2015_CVPR} proposed a learning-based approach to synthesize diverse view variations. This method is capable of rendering different models of inputs.
Zhou \etal~\cite{zhou2016view} proposed appearance flow method to synthesize object with extreme view variations, but this method is limited to a single object with a homogeneous background.
These existing studies handled objects with limited shape variances and inward-looking view, while the proposed work handles outward-looking views with relatively diverse scenes.

\vspace{-2mm}
\subsection{Beyond Camera Viewpoint}
Framebreak~\cite{framebreak} yielded impressive results in generating partial panorama from images with a small FOV.
The method requires the manual selection of reference images to be aligned with input images.
Guided patch-based texture synthesis is used to generate missing pixels.
The process requires reference images with high similarity with the input.

Xiao~\etal~\cite{SUN360} predicted the viewpoint of a given panoramic observation.
The prediction generated rough panorama structure and compass-like prediction in determining the location of the viewpoint in 360{\degree}.
Georgoulis~\etal~\cite{Georgoulis_CAMERA_2017} estimated environmental map from the reflectance properties of input images by utilizing these properties from a foreground object to estimate the background environment in a panoramic representation.
By contrast, the proposed method does not rely on reference images as the input, and it synthesizes actual panoramic imagery.

\vspace{-1mm}

\subsection{Generative Models}
Generative models enjoy tremendous success in the image synthesis domain.
The generative adversarial network (GAN)~\cite{NIPS2014_GAN} synthesize images from noise instance and works well on images with low resolutions but mainly struggles on images with high-resolution and suffers from instability problem during training.
During the last few years, several architectures and variants~\cite{wgan_arjovsky2017, Lucic2017AreGC, Mao2017LeastSG, pix2pix2017, wang2018pix2pixHD, karras2018progressive} have been proposed to overcome these major limitations and improve the results.
Following the success of these models, recent image inpainting based on generative models~\cite{IizukaSIGGRAPH2017, Pathak_2016_CVPR, Yeh_2017_CVPR} are widely utilized to fill in the missing regions.
Wang~\etal~\cite{Wang_2019_CVPR} handled the outpainting task by generating full images from smaller input by propagating the learned features from the small size images.


\begin{figure*}[t]
	\begin{center}
		\includegraphics[width=\linewidth]{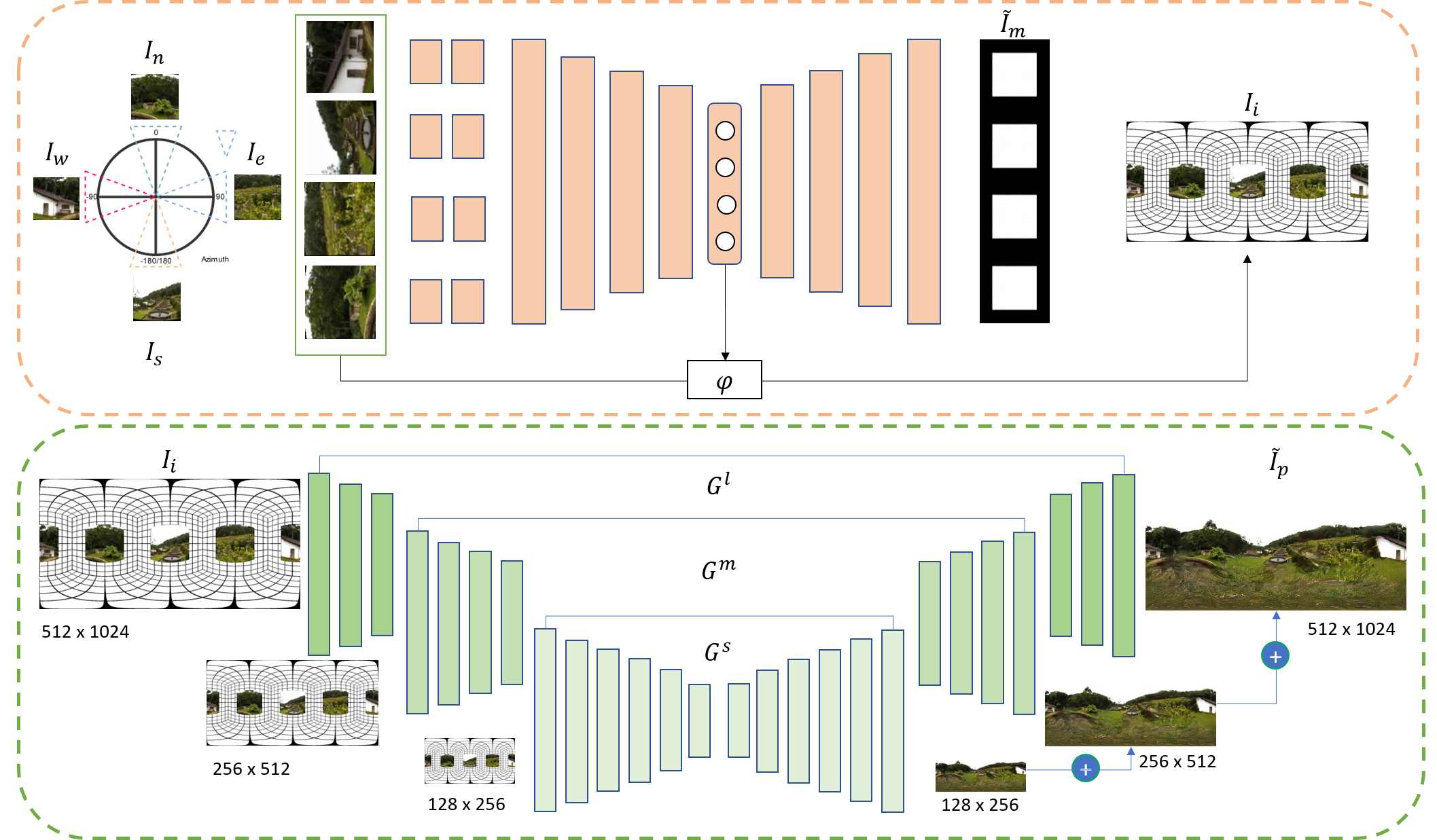}
	\end{center}
	\vspace*{-3mm}
	\caption{The architecture of the proposed network.
		We show the visualization of the relative FOV estimation network (top) and panorama synthesis network (bottom).
		The network takes sequence of 4 input images to estimate the equirectangular panorama with missing pixels.
		This equirectangular panorama is then used as the input for the panorama synthesis network.
		The synthesis process is done on each scale on small, medium, and large.
	}
	\label{fig:proposed_framework}
\end{figure*}

\section{Proposed Approach}
\begin{figure}[t]
	\begin{center}
		\includegraphics[width=\linewidth]{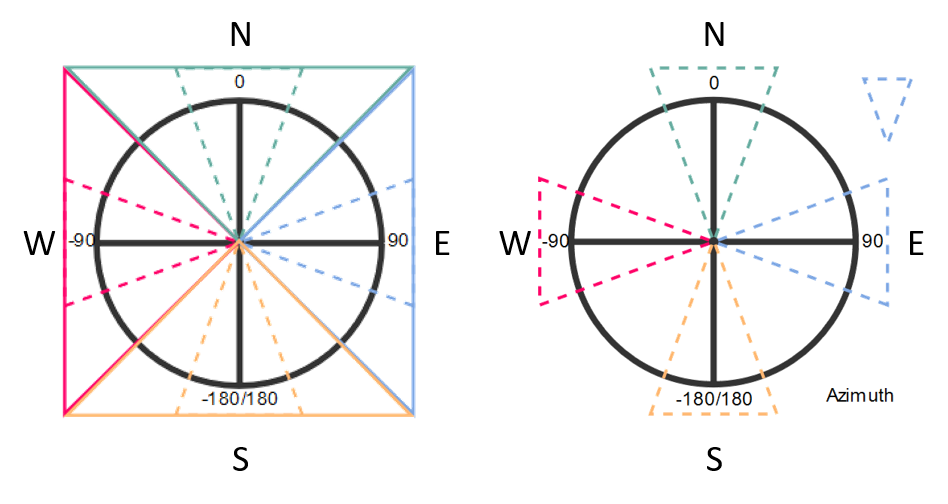}
	\end{center}
	\vspace*{-5mm}
	\caption{
		{FOV visualization}. The triangles are observation FOV looking from the vertical center viewpoint.
		Solid triangles are observation with  90{\degree} FOV.
		Dashed triangles are observations with less than 90{\degree} FOV which form partial panorama.
	}
	\label{fig:compass_rose}
	\vspace*{-5mm}
\end{figure}

We designed and utilized a CNN with GAN-based framework to address the estimation and synthesis problem, as shown in Figure~\ref{fig:proposed_framework}.
The input to the proposed network is an ordered sequence of 4 images.
Each observation is performed onto 4 cardinal directions of the compass rose: north, west, south, and east, as shown in Figure~\ref{fig:compass_rose}.\footnote{NWSE direction is just for explanation and can be a `random' 4 directions as long as they are roughly 90{\degree} apart.}
The viewing directions of 4 inputs only need to be roughly perpendicular.
As an ideal case, if 4 images are captured to each perpendicular direction with a 90{\degree} FOV and 1:1 aspect ratio, a complete horizontal panorama can be formed simply by concatenating them together.

\begin{figure*}[t]
	\begin{center}
		\begin{subfigure}{0.185\linewidth}
			\includegraphics[width=\linewidth]{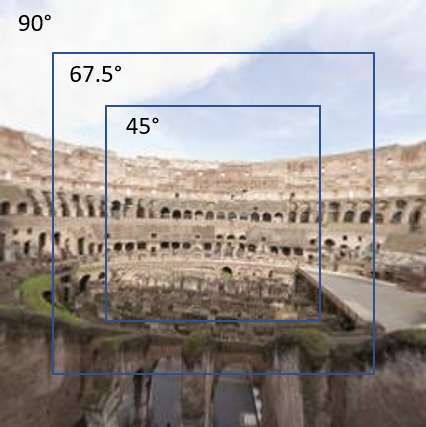}
			\caption{}
		\end{subfigure}
		\begin{subfigure}{0.185\linewidth}
			\includegraphics[width=\linewidth]{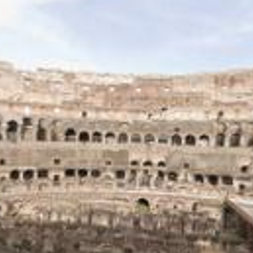}
			\caption{}
		\end{subfigure}
		\begin{subfigure}{0.185\linewidth}
			\includegraphics[width=\linewidth]{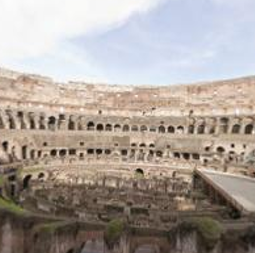}
			\caption{}
		\end{subfigure}
		\begin{subfigure}{0.185\linewidth}
			\includegraphics[width=\linewidth]{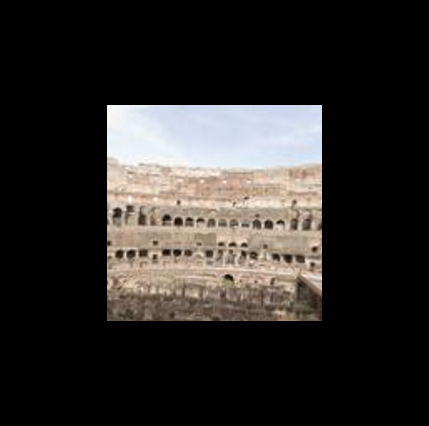}
			\caption{}
		\end{subfigure}
		\begin{subfigure}{0.185\linewidth}
			\includegraphics[width=\linewidth]{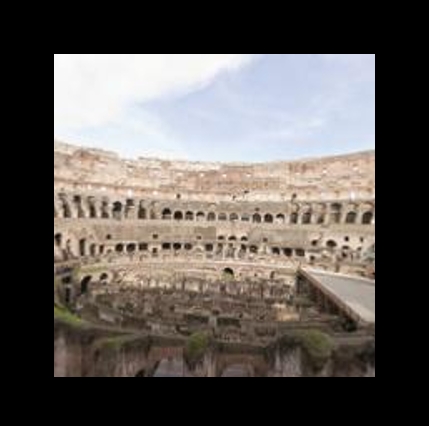}
			\caption{}
		\end{subfigure}
		
	\end{center}
	\vspace*{-6mm}
	\caption{The meaning of relative FOV. (a) shows image with 90{\degree} FOV.
		The blue line shows scene coverage of different FOV. (b) and (c) shows
		images at captured in 45{\degree} and 67.5{\degree} FOV. The outputs of our
		FOV estimation network are shown in (d) and (e). The black region is the
		empty pixel, and image in the center is rescaled to match the panorama FOV. }
	\label{fig:fov}
	\vspace*{-4mm}
\end{figure*}

\subsection{Relative FOV Estimation}
We define the scene in 4 cardinal directions as $I = [I_n, I_w, I_s, I_e]$ for the northern, western, eastern and southern direction, respectively.
Images with smaller FOVs only present a smaller portion of the scenes, as illustrated in Figure~\ref{fig:fov}.
The typical scene captured with standard camera normally have a FOV less than 90{\degree}.
As a result, they do not form any connection or overlapping when concatenated.
$I$ forms disconnected partial panorama on the horizontal axis and is used as the input.

CNN architecture is utilized to solve the FOV estimation task.
Images from the same scene captured with 90{\degree} FOV are shown in Figure~\ref{fig:fov}(a) and smaller FOV in Figure~\ref{fig:fov}(b).
The network takes smaller FOV images as the inputs and estimates the relative FOV (Figure~\ref{fig:fov}(d)).
The input is processed on the multi-stream layers before joining them together, followed by encoder and decoder structure.
In the bottleneck layer, the network outputs nodes representing the relative FOV angle, which is treated in a similar manner with classification task.
Softmax cross-entropy loss is used as the objective function for the classification task,
\begin{equation}
H_{\phi} = - \sum_{i}^N \{{\tilde{y}_i \log(y_i) + (1-\tilde{y}_i) \log (1-y_i)}\},
\end{equation}
where $\tilde{y}$ is the predicted FOV angle and $y$ is the ground-truth. 
Note that this approach does not estimate the actual FOV.
The angle of the FOV is relative to the size of the images, thus by rescaling these images, the estimated relative FOV which corresponds to each viewing directions can be obtained.
The main goal of this network is to estimate the relative FOV served as a constraint to help the synthesis process.

The decoder layers synthesize mask-like image structure in horizontal panorama format $\tilde{I}_{mask}$.
The purpose of the mask image synthesis is solely for guiding the FOV estimation.
We used L1 loss objective function for the image structure synthesis, defined as follows,
\begin{equation}
L_{mask} = {||I_{mask} - \tilde{I}_{mask}||_1},
\end{equation}
where $I_{mask}$ is the ground truth horizontal panorama mask.
The full objective function of the relative FOV estimation is defined as,
\begin{equation}
L_{fov} = H_{\phi} + \lambda_{mask} L_{mask},
\end{equation}
We processed the input images using the estimated relative FOV angle $\phi$ by rescaling the image size and adding the zero padding.
Before the input is fed into the panorama synthesis stage, the processed input is warped to equirectangular format.
The final output from this estimation stage is formulated as,
\begin{equation}
I_{i} = M(p(I,\phi))
\end{equation}
where $p(\cdot)$ is the scaling and padding function and $M(\cdot)$ is the warping function from horizontal panorama to an equirectangular panorama. $I_{i}$ is the equirectangular panorama with missing pixel region.


\subsection{Panorama Synthesis}
The panorama synthesis problem is treated with a hierarchical approach.
Images with high resolution have more information distribution in space than images with lower resolution, making the training progress difficult.
Instead of aiming for the global minimum in a single run, we enforced the network to learn step by step in achieving the local minimum on each hierarchy.
This step can be regarded as providing soft guidance to help the network converge.

Similar to the image pyramid, the task is decomposed into synthesizing images with different scales.
To this end, we decomposed the input images into three hierarchies, namely, small, medium, and large scale.
The large scale is the target panorama with $512\times1024$ resolution.
The medium scale is downsampled by the scale factor of 2 from the original scale with $256\times512$ resolution.
The small scale is downsampled again by the scale factor of 2 with $128\times256$ resolution.
\vspace*{-3mm}

\subsubsection{Unified Generator}
The generator consists of three sub-networks, namely $G^s$, $G^m$, and $G^l$, each corresponding to a different scale.
These sub-networks contain both the input-output bridge.
The input bridge is used to facilitate the connection on a different scale, while the output bridge is used to map the output from high dimensional channels to RGB channels.
The smallest scale generator is utilized as the base generator for the entire training process.

The training process is performed separately on each scale starting from the smallest scale.
To train the next hierarchy, the base generator $G^s$ is unified with the generator $G^m$.
Weight parameters learned from the previous scale are reused and fine-tuned at each increased scale.
This training process is repeated until the original scale is reached.
As a result, the generator $G^l$ is unified with both $G^m$ and $G^s$ at the largest scale to form a single generator.

We proposed short- and long-term connection for the network architecture.
The short-term connection is employed with residual blocks~\cite{He2016_CVPR} in the base generator and the input bridge for the large scale images.
The long-term connection is used to maintain the feature connectivity between layers.
Unlike~\cite{SAGAN} where the attention map is used in its current layer, the attention map is propagated by concatenating them from early blocks with last blocks as the long-term connection.

\subsubsection{Multiscale Residual}
The base generator takes an input from small-scale image and outputs small-scale panorama.
For the medium-scale, this output is used as the residual by performing upsampling operation and is added to the medium-scale panorama output subsequently.
The same rule follows for training the large-scale images by keeping the input bridge of the small- and medium-scale images, followed by the residual from the output bridge.
The function is defined as,
\begin{align}\label{Eq:5}
\hat{I}^{l}_p = G^{l}({I}^{l}_i) + f(\hat{I}^{m}_p), \\
\hat{I}^{m}_p = G^{m}({I}^{m}_i) + f(\hat{I}^{s}_p), \\
\hat{I}^{s}_p = G^{s}({I}^{s}_i),
\end{align}
where $\hat{I}^{*}_p$ is the panorama output and $I^{*}_i$ is the input at each scale.
$G^{*}$ and $f$ denote the network function and the upsampling operation, respectively.

\subsubsection{Network Loss}
For each image scale, multiple patch discriminators are employed, that is $D^s$, $D^m$, and $D^l$.
They have identical architecture with encoder-like structure across all scales.
At a small scale, only single discriminator $D^s$ is used.
Both $D^s$ and $D^m$ are used at the medium scale and use all three discriminators at the large scale.
For the medium and large scales, the output images from the current resolution are downscaled to match the resolution at each scale.

Conditional GAN has a discriminator network $D$ which takes both input and output from the generator.
We utilized the conditional architecture with LSGAN~\cite{LSGAN} loss as the adversarial loss for the synthesis problem.
The objective function for the adversarial loss is defined as,
\begin{multline} \label{Eq:1}
L_{adv}^{\dot{*}}(G,D) = \frac{1}{2}\mathbf{E}_{I_{p}}[(D(I_{i},I_{p}) - 1)^2] + \\
\frac{1}{2}\mathbf{E}_{I_{i},\hat{I}_{p}}[(D(I_{i},\hat{I}_{p]}))^2],
\end{multline}
where $L_{adv}{\dot{*}}$ denotes the adversarial loss at a specific scale.
The independent loss function at the largest scale is defined as
\begin{equation}
L_{adv}^{l} =  L_{adv}^{\dot{s}} +  L_{adv}^{\dot{m}} + L_{adv}^{\dot{l}},
\end{equation}
where $L_{adv}^{*}$ denotes the total adversarial loss from the discriminators.

Pixel loss is employed to supervised loss to facilitate the synthesis task.
We used L1 loss to minimize the generated output with the ground truth panorama.
The network objective function is defined as,
\begin{equation}
L_{pix}^{*} = \mathbf{E}_{I_{p},\hat{I}_{p}}[||I_{p}^{*} - \hat{I}_{p}^{*}||_1] \label{Eq:2},
\end{equation}
where $I_{i}$ and $\hat{I}_{p}$ are the input image and the output from the generator $G(I_{i})$, respectively.
The pixel loss is defined as the $L1$ distance between the ground truth panorama
$I_{p}$ and $\hat{I}_{p}$.
To further improve the realism of the generated output, we added the perceptual loss obtained from the pre-trained weight of VGG networks, which are defined as,
\begin{equation}
L_{vgg}^{*} = \sum_{i} X_{i}(I_{p}^{*}) - X_{i}(\hat{I}_{p}^{*}),
\end{equation}
where $X$ is the extracted i-th features from the VGG network.
The overall loss for the network is defined as,
\begin{equation}\label{Eq:3}
L^{*} = {arg} \mathop{min}_G \mathop{max}_D \ {L_{adv}^{*}} + {\lambda} {L_{pix}^{*}} + {\lambda} {L_{vgg}^{*}},
\end{equation}
which is the total adversarial GAN loss, with the pixel loss scaled by $\lambda$ constant factor.
\section{Experimental Results}
We encourage the readers to refer to the supplementary material for more results on general scene, different input setup, and the visualization of the panorama's free view point video.
The training procedure is done separately on the field of view estimation network and panorama synthesis network.
Our networks are built on convolutional blocks followed with instance normalization and leaky ReLU activation.
The input bridge maps three RGB channels onto 64-dimensional layers.
The output bridge maps 64-dimensional channels back to a RGB channels.
ADAM~\cite{Kingma2014AdamAM} optimizer is used with learning rate $\alpha = 0.0002$, $\beta_{1} = 0.5$, and $\beta_{2} = 0.99$.
The input and output are normalized to [$-1,1$].

\begin{figure*}
	\begin{center}
		
		{\includegraphics[width=0.234\linewidth]{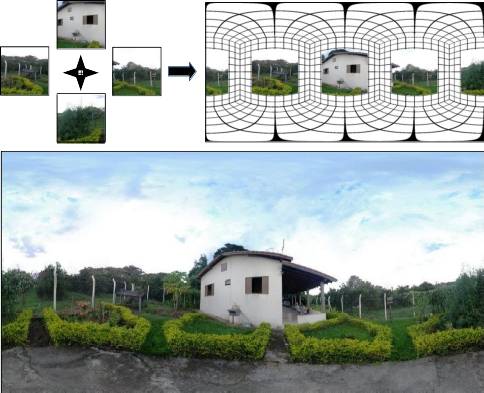}}~
		{\includegraphics[width=0.375\linewidth]{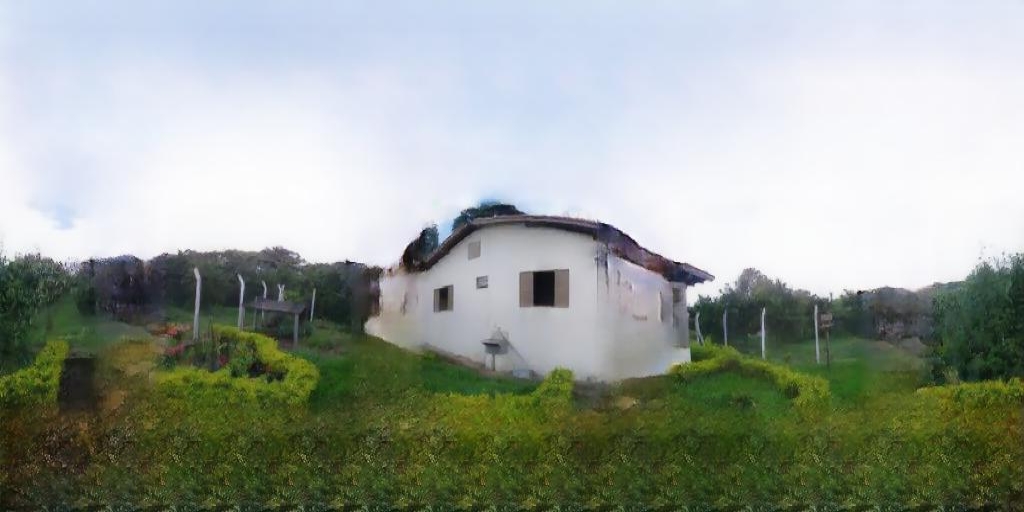}}~
		{\includegraphics[width=0.375\linewidth]{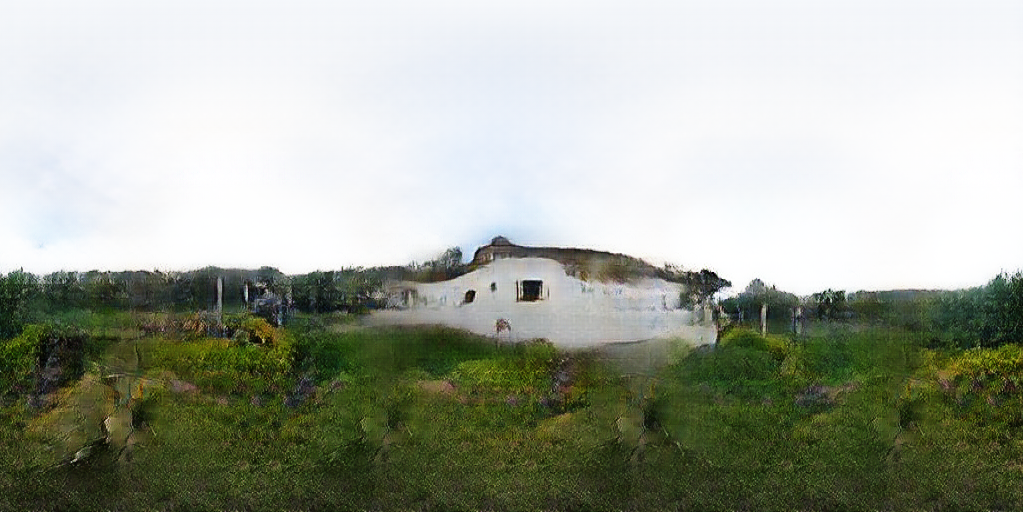}}~
		\vspace{2.0mm}		
		
		\stackunder[10pt]{\includegraphics[width=0.234\linewidth]{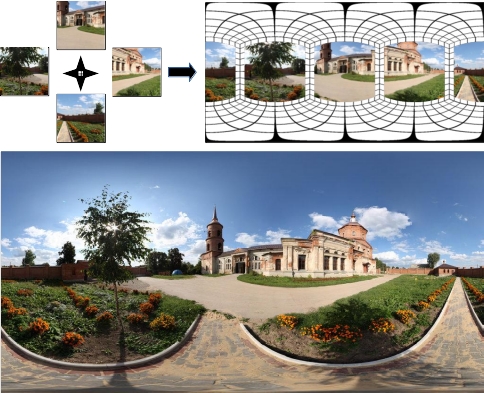}}
		{(a)}~
		\stackunder[10pt]{\includegraphics[width=0.375\linewidth]{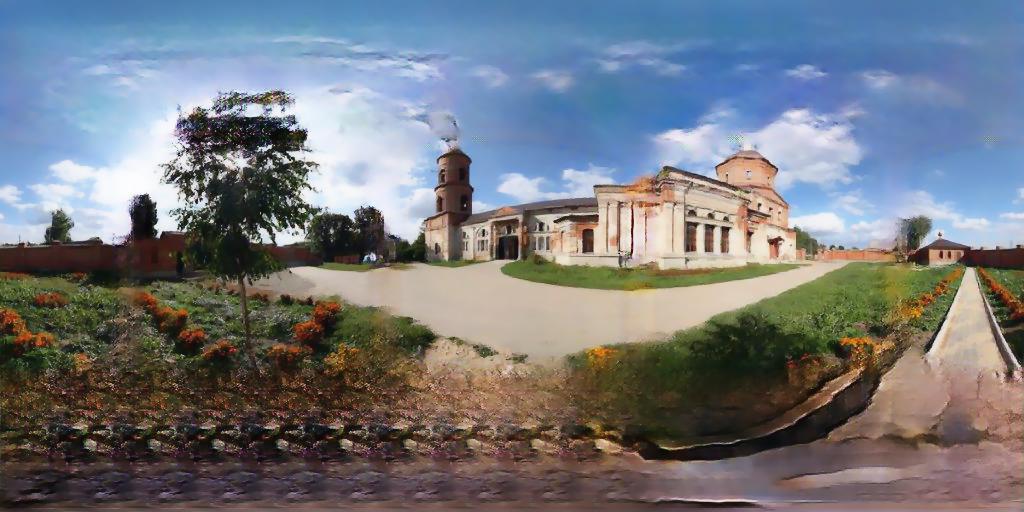}}
		{(b)}~
		\stackunder[10pt]{\includegraphics[width=0.375\linewidth]{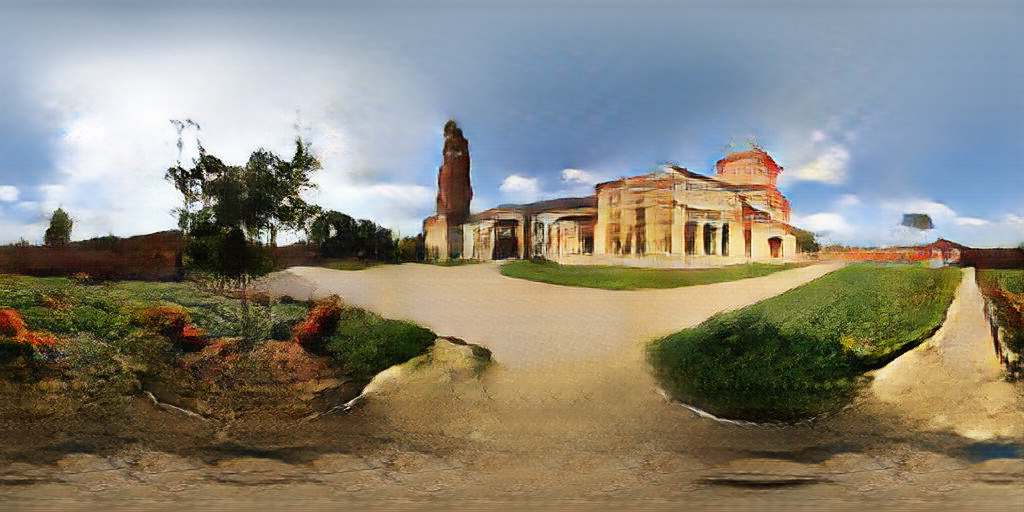}}
		{(c)}
	\end{center}
	\vspace*{-4mm}
	\caption{{Synthesized panoramas in 512$\times$1024}.
		The input of 4 cardinal direction shown in upper left (a).
		The relative FOV is estimated and warped into partial panorama in upper right (a).
		Ground truth is shown in bottom (a). We visualized our synthesized 360{\degree} panoramas in (b).
		The results are compared with pix2pixHD~\cite{wang2018pix2pixHD} in (c).
		The proposed method produces sharper panorama images while the baseline work produces smoother and blurrier results.}
	\label{fig:result}
\end{figure*}



\subsection{Dataset}
For the training and evaluation of the proposed network, outdoor panoramic images are used from the SUN360~\cite{SUN360} dataset.
The dataset was split into a training set (80\%) and a testing set (20\%).
The dataset contains approximately 40,000 images (512$\times$1024) with various scenes rendered in equirectangular format.
To accommodate the data with the proposed framework, we rendered the panorama to a cube map format by warping it to the planar plane. The viewing direction is split into the 4 horizontal sides of the cube.
The FOV angle is rendered randomly from 45{\degree} to 75{\degree} with identical angle on each NWSE direction.
The vertical center is located at 0{\degree}, and the horizontal center is located at 0{\degree}, 90{\degree} 180{\degree}, and 270{\degree}.

\begin{figure*}[t]
	\begin{center}
		{\includegraphics[width=0.234\linewidth]{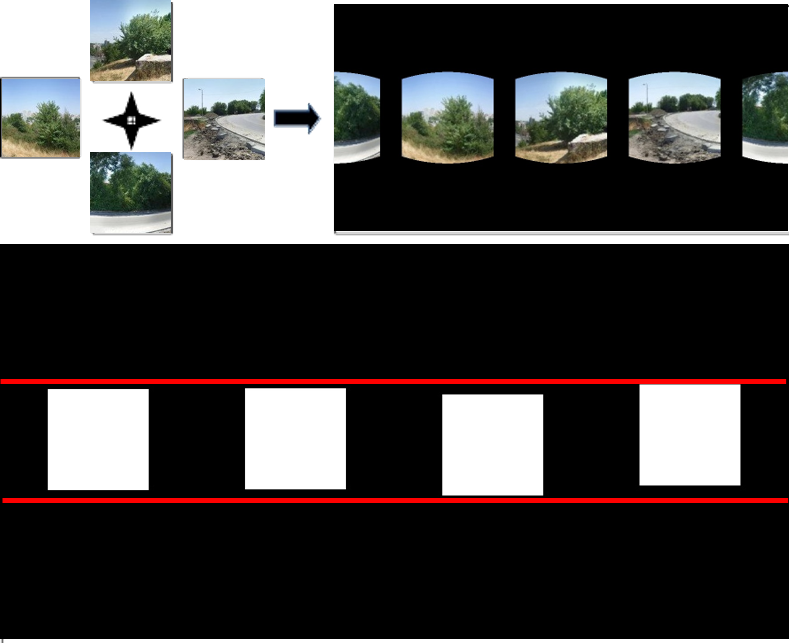}}~
		{\includegraphics[width=0.375\linewidth]{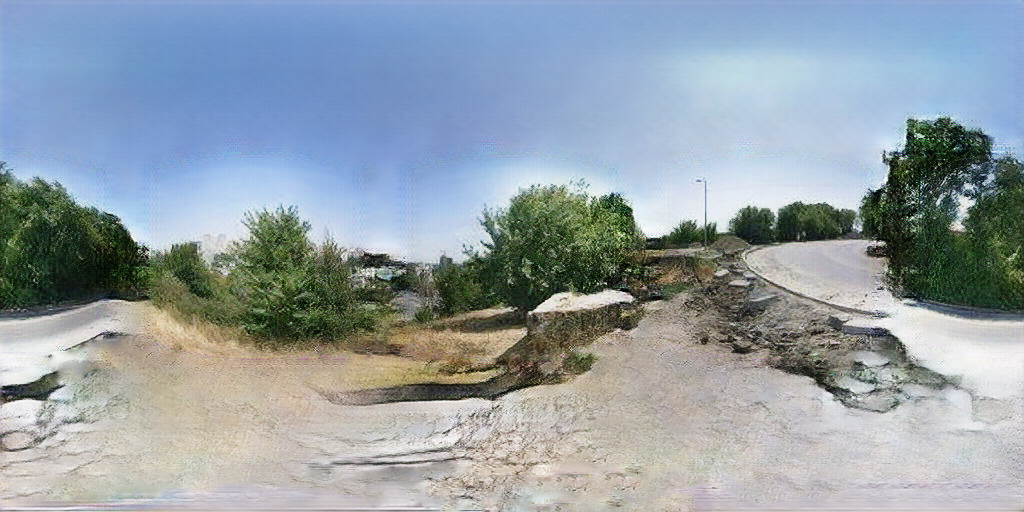}}~
		{\includegraphics[width=0.375\linewidth]{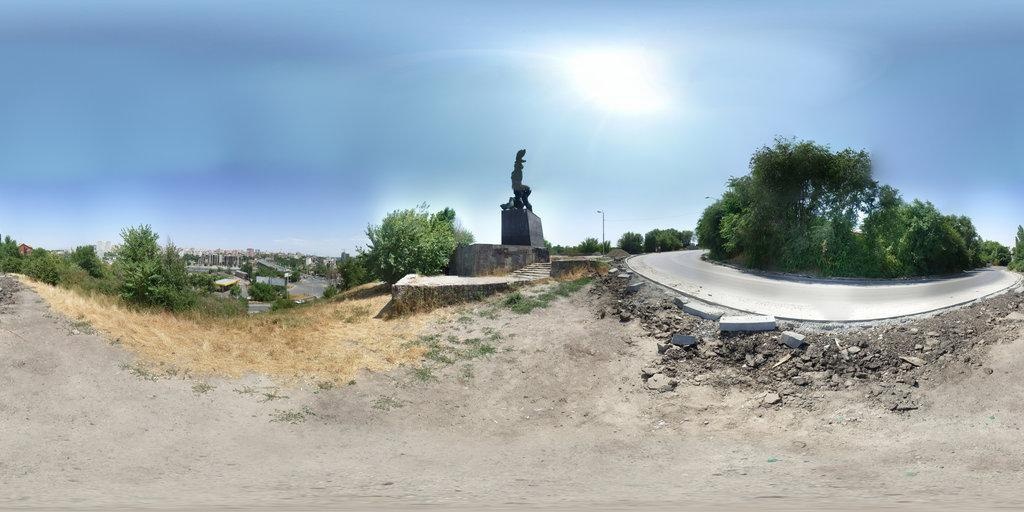}}
		\vspace*{-2mm}\\

		\stackunder[10pt]{\includegraphics[width=0.234\linewidth]{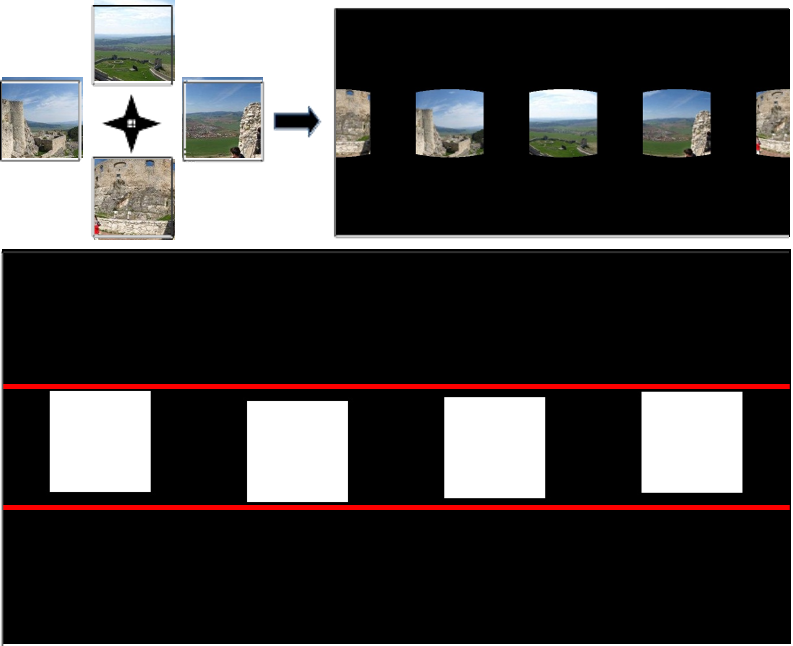}}
		{(a)}~
		\stackunder[10pt]{\includegraphics[width=0.375\linewidth]{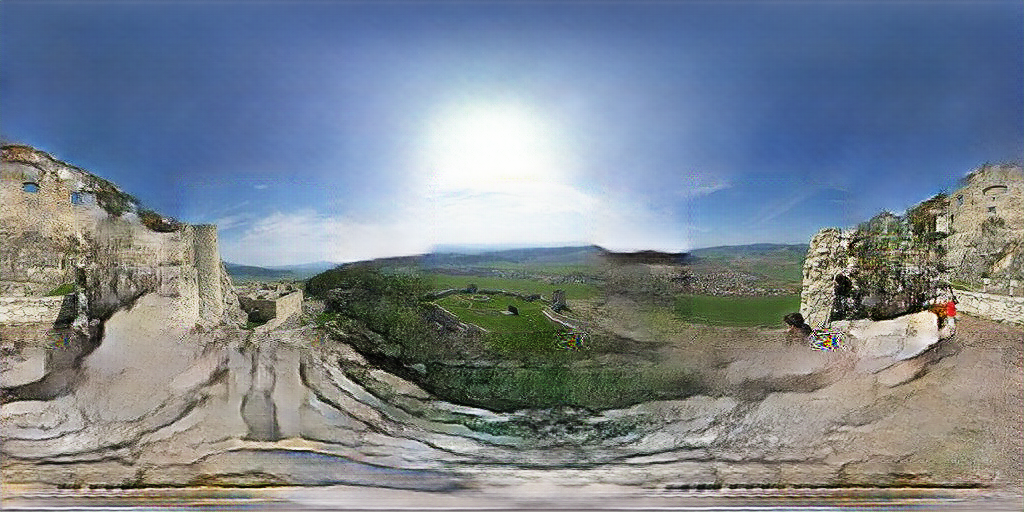}}
		{(b)}~
		\stackunder[10pt]{\includegraphics[width=0.375\linewidth]{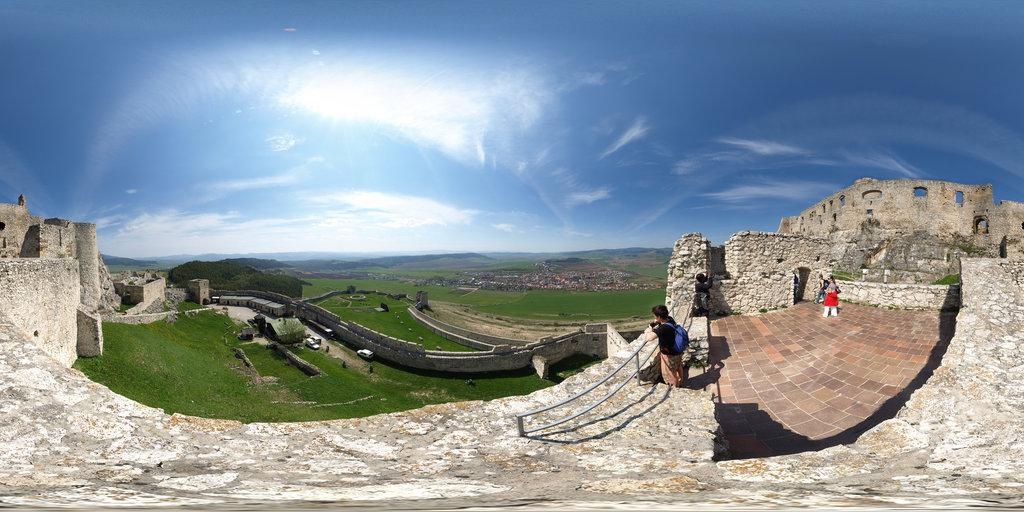}}
		{(c)}~
	\end{center}
	\vspace*{-6mm}
	\caption{Synthesized panorama in 512$\times$1024 from non-horizontal input. The input of 4 cardinal direction shown in upper left (a).
		The relative FOV is estimated and warped into partial panorama in upper right (a).
		Bottom (a) shows a visualization of the non-horizontal input.
		The output and ground truths are shown in (b) and (c).}
	\label{fig:result_ovl}
	\vspace*{-2mm}
\end{figure*}

\begin{figure*}[t]
	\begin{center}
		\includegraphics[width=0.234\linewidth]{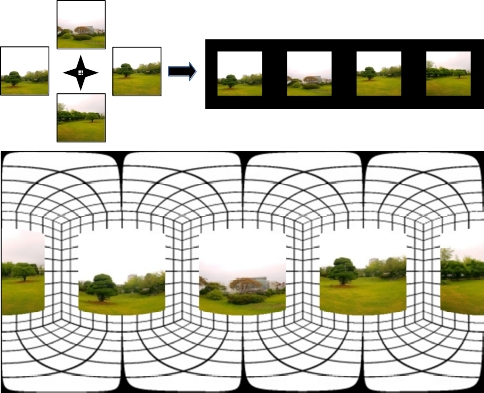}~
		\includegraphics[width=0.375\linewidth]{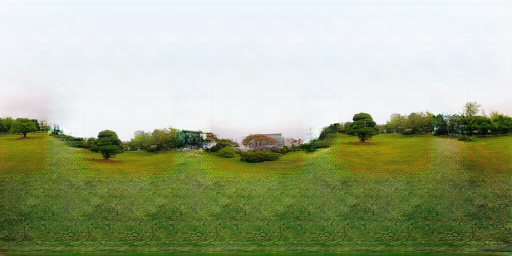}~
		\includegraphics[width=0.375\linewidth]{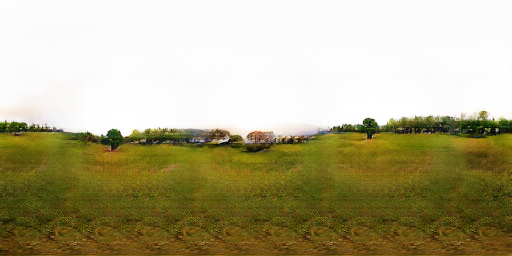}~\\
		\vspace{1mm}
		
		\stackunder[5pt]{\includegraphics[width=0.234\linewidth]{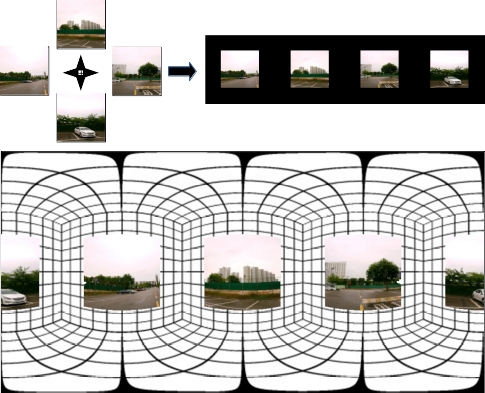}}
		{(a)}~
		\stackunder[5pt]{\includegraphics[width=0.375\linewidth]{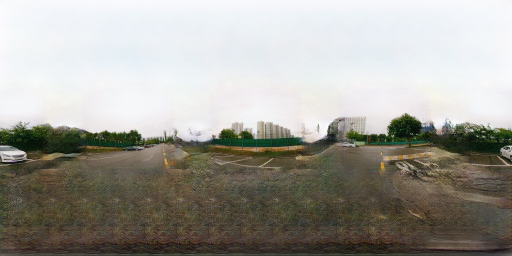}}
		{(b)}~
		\stackunder[5pt]{\includegraphics[width=0.375\linewidth]{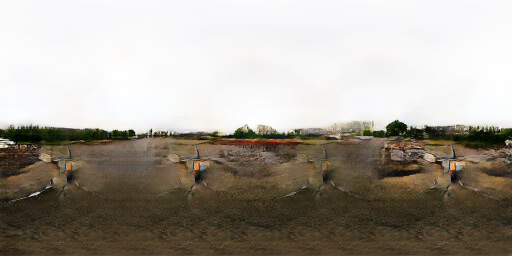}}
		{(c)}~\\

	\end{center}
	\vspace*{-6mm}
	\caption{{Results of smartphone images}. The input of 4 cardinal direction shown in upper left (a).
		The relative FOV is estimated and warped into partial panorama in upper right and bottom (a).
		We visualize our synthesized 360{\degree} panoramas in (b).
		The results are compared with pix2pixHD~\cite{wang2018pix2pixHD} in (c).}
	\label{fig:real_result}
	\vspace*{-3mm}
\end{figure*}

\subsection{Qualitative Evaluation}
High resolution synthesis using conditional GAN is not widely studied.
pix2pixHD~\cite{wang2018pix2pixHD} is mainly presented to synthesize images from semantic labels.
However, the conditional properties of GAN in the framework can be used as
a baseline approach in our case.
The proposed method clearly outperforms the baseline, as shown in Figure~\ref{fig:result}.
Note that the repetitive pattern at the bottom is due to the training data which contains circular shape of watermark in it.
Additional result on the case where the input is non-horizontally aligned is shown in Figure~\ref{fig:result_ovl}.

The outdoor scene dataset also exhibits greater variance than the dataset of faces or road scenes in semantic-to-image synthesis which makes the task more challenging.
The typical results from the baseline are synthesized smoothly, but they appear cartoonish and lacking in detail, whereas our result is relatively sharp.
We believe that this difference is achieved by hierarchical synthesis method that we proposed.

\subsection{Quantitative Evaluation}
\vspace*{-1.5mm}
We measured the accuracy and the percent error of the estimated FOV in Table~\ref{Table:quantitative_result1}.
The error shows how different the predicted value from the actual ground truth.
It is shown that the proposed method yields better accuracy with very low error.
The low error denotes that the FOV angle is off by only a few degrees.
The FOV estimation with pixel synthesis can only generate rough image structure and presents several improvements over the baseline, but have lower accuracy compared to mask structure synthesis.

\renewcommand{\arraystretch}{1.1}
\begin{table}
	\centering
	\begin{tabu}  {  X[c]  X[c]  X[c]}
		\hline
		\textbf{Method}                  & \textbf{Accuracy}     & \textbf{Error}\\ \hline
		DeepFocal~\cite{deepfocal2015}   & {0.56}                & {0.33}            \\
		Ours (w/ mask)                   & {0.78}                & {0.05}            \\
		Ours (w/ pixel)                  & {0.76}                & {0.06}            \\
		\hline
	\end{tabu}
	\vspace*{-2mm}
	\caption{Accuracy and error of the FOV estimation.}
	\label{Table:quantitative_result1}
	\vspace*{-5mm}
\end{table}
Generative model's evaluation metrics are not well established. Each metrics has advantages and disadvantages.
Several works on super resolution~\cite{Ledig_2017_CVPR} and image synthesis~\cite{Regmi_2018_CVPR} using the generative model employ structural-similarity~(SSIM) and peak signal to noise ratio (PSNR).
For the sake of comparison, we evaluated the proposed method on both SSIM and PSNR.
The proposed method yields better average SSIM and PSNR over the baseline, shown in Table~\ref{Table:quantitative_result2}.

We further evaluated the performance of the proposed method with a user study.
The study is conducted with 50 blind pairwise comparisons on 20 participants.
For the high resolution output, our result is preferred by 89\% of users compared to the pix2pixHD~\cite{wang2018pix2pixHD}.
For the small resolution output trained with random mask, our result is preferred by 92\% of users compared to SRN~\cite{Wang_2019_CVPR}.

\renewcommand{\arraystretch}{1.1}
\begin{table*}[t]
	\centering
	\begin{tabu}  {  X[c]  X[c]  X[c] X[c] X[c] X[c] X[c]}
		\hline
		\textbf{Metric}  & \textbf{Proposed}  & \textbf{Proposed-NH}& \textbf{Proposed-NC}  & \textbf{Proposed-SD}  & \textbf{pix2pixHD~\cite{wang2018pix2pixHD}}\\ \hline
		SSIM             & {0.4528 }      & {0.3951}     & {0.4452}    & {0.4312}    & {0.3921}            \\
		PSNR             & {15.971}       & {15.273}     & {15.927}    & {15.336}    & {15.264}            \\
		\hline
	\end{tabu}
	\vspace*{-2mm}
	\caption{Quantitative comparison with SSIM and PSNR (in dB) scores between the proposed method and the baseline. The variant of proposed method is included without hierarchical structure (NH), without long-term connection (NC), and with a single discriminator (SD).}
	\label{Table:quantitative_result2}
	\vspace*{-2mm}
\end{table*}


\subsection{Real-World Data in the Wild}
\vspace*{-1.5mm}
Furthermore, we showed additional results for real-world scenes taken with a smartphone, as shown in Figure~\ref{fig:real_result}.
Our capturing process is convenient as it only requires a few seconds to capture 4 images in different view directions instead of capturing images continuously.
The input images are captured with a rough estimation of 4 cardinal directions.
There could be inaccuracy where the viewing directions are not perpendicular to each other.
However, a visually plausible result can still be generated.

\subsection{Ablation Study}

\begin{figure*}[t]
	\begin{center}
		\stackunder[5pt]{\includegraphics[width=0.245\linewidth]{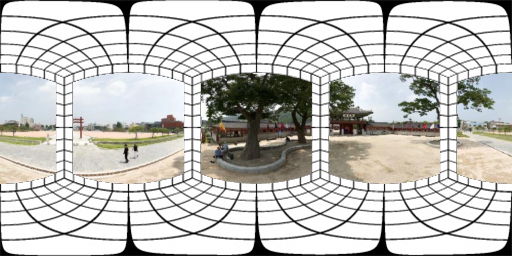}}
		{(a)}~
		\stackunder[5pt]{\includegraphics[width=0.245\linewidth]{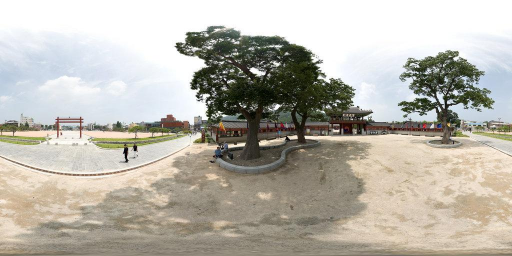}}
		{(b)}~
		\stackunder[5pt]{\includegraphics[width=0.245\linewidth]{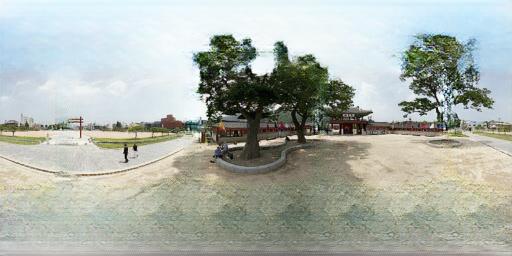}}
		{(c)}~
		\stackunder[5pt]{\includegraphics[width=0.245\linewidth]{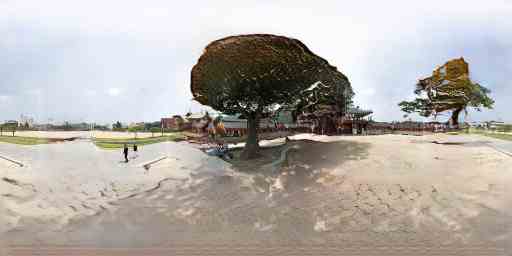}}
		{(d)}
	\end{center}
	\vspace*{-6mm}
	\caption{Ablation study: The effect of hierarchical synthesis. (a) Input, (b) Ground truth, (c) Hierarchical synthesis, (d) Direct synthesis.}
	\label{fig:abla_hierarchy}
	\vspace*{-3mm}
\end{figure*}

\paragraph{Hierarchical structure (NH)}
We conducted an experiment by training the network in two different manners, namely the direct and hierarchical synthesis.
The compared network is built on a similar design without any changes but trained without a hierarchical structure.
The output is shown in Figure~\ref{fig:abla_hierarchy}.
By learning from the lower scale images, the output trained in a hierarchical manner is synthesized with better details compared with direct synthesis.

\begin{figure*}[!t]
	\begin{center}
		\stackunder[5pt]{\includegraphics[width=0.245\linewidth]{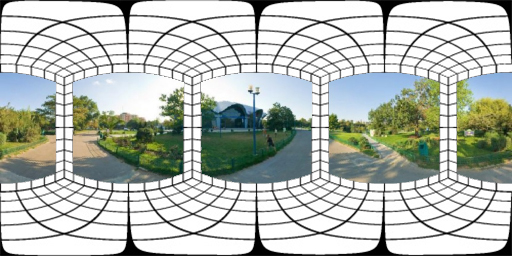}}
		{(a)}~
		\stackunder[5pt]{\includegraphics[width=0.245\linewidth]{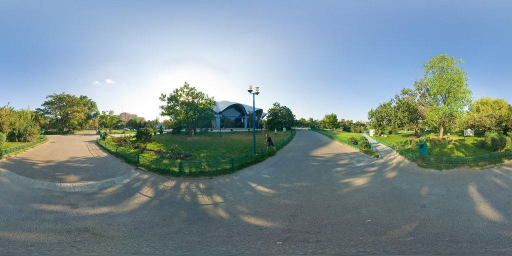}}
		{(b)}~
		\stackunder[5pt]{\includegraphics[width=0.245\linewidth]{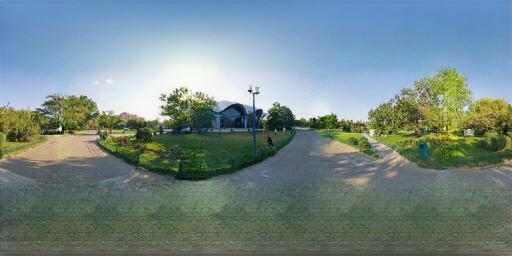}}
		{(c)}~
		\stackunder[5pt]{\includegraphics[width=0.245\linewidth]{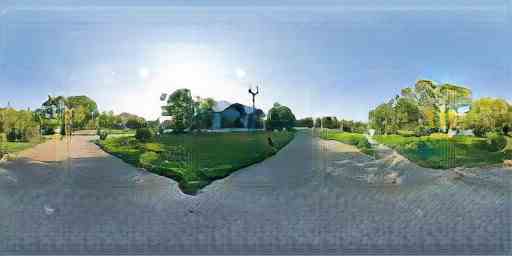}}
		{(d)}\\
	\end{center}
	\vspace*{-6mm}
	\caption{Ablation study: The effect of long-term connection. (a) Input, (b) Ground truth, (c) With long-term connection, (d) Without long-term connection.}
	\label{fig:long_connect}
	\vspace*{-3mm}
\end{figure*}

\begin{figure*}[!t]
	\begin{center}
		\stackunder[5pt]{\includegraphics[width=0.245\linewidth]{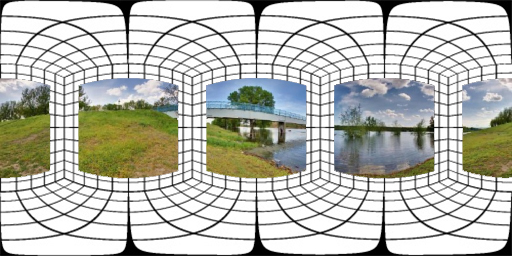}}
		{(a)}~
		\stackunder[5pt]{\includegraphics[width=0.245\linewidth]{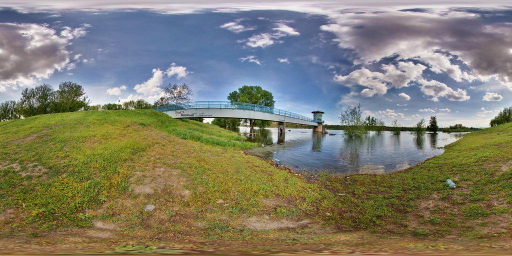}}
		{(b)}~
		\stackunder[5pt]{\includegraphics[width=0.245\linewidth]{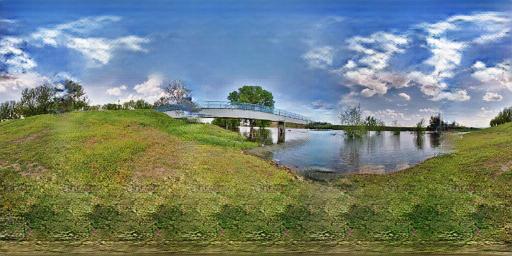}}
		{(c)}~
		\stackunder[5pt]{\includegraphics[width=0.245\linewidth]{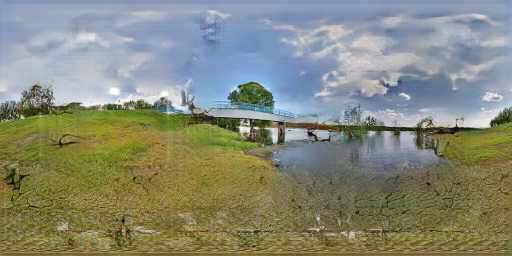}}
		{(d)}\\
	\end{center}
	\vspace*{-6mm}
	\caption{Ablation study: The effect of different discriminators. (a) Input, (b) Ground truth, (c) Multi discriminators, (d) Single discriminator.}
	\label{fig:multi_disc}
	\vspace*{-5mm}
\end{figure*}

\vspace*{-4mm}
\paragraph{Short- and long-term connections (NC)}
We investigated the effect of using the long-term connection on the training, as shown in Figure~\ref{fig:long_connect}.
The long-term connection acts like the prior input for the network.
Features from early convolutional blocks encode similar properties of the input which is particularly useful for our case where the network has a deep architecture.
Early features help guide the network to maintain similar properties between the input and output.

\vspace*{-4mm}
\paragraph{Multiple discriminators (SD)}
During the experiments, we found that using multiple discriminators can stabilize the training process.
The comparison between them is shown in Figure~\ref{fig:multi_disc}.
Output trained with multiple discriminators produces better output with less visible artifacts.

Quantitative result on all type of the ablation studies is evaluated in the Table~\ref{Table:quantitative_result2}.

\subsection{Limitations}
\vspace*{-1.5mm}
Although the performance of the proposed method is promising, it has a few limitations.
First, the proposed method struggles in synthesizing scene with highly complex structure with many trees, foliage, and people.
Second, the FOV estimation network is limited to handle input in an ordered non-overlapping sequence with an identical FOV angle.
Third, in order to use the synthesized panorama in virtual reality equipment, the resolution should be much higher than current maximum resolution (512$\times$1024).
We hope that those limitations can be handled properly in the further research.
\vspace*{-3.mm}
\section{Conclusion}
\vspace*{-2.5mm}
In this study, the novel method is presented to synthesize $360\degree\times180\degree$ panorama from a sequence of wide baseline partial images.
The proposed method generated high-resolution panoramic images by estimating the FOV and hierarchically synthesizing panorama.
Experimental results showed that the proposed method produced 360{\degree} panorama with good quality.
Furthermore, it outperformed the conventional method and extendable to non-panorama scenes and images captured by a smartphone camera.
\vspace*{-3.0mm}

\section*{Acknowledgement}
\vspace*{-1.5mm}
This work was supported by Samsung Research Funding Center of Samsung Electronics under Project Number SRFC-IT1702-06.

{\small
\bibliographystyle{ieee}
\bibliography{egbib}
}

\end{document}